%% file: Multiclass.tex
\documentclass[letterpaper, 10 pt, conference]{ieeeconf}

\IEEEoverridecommandlockouts  

\input{preamble/common}

\title{\LARGE \bf
Multi-class Temporal Logic Neural Networks
}

\author{Danyang Li, Roberto Tron
\thanks{$^{1}$This work was supported in part by MIT Lincoln Laboratory MAESTRO program.}
\thanks{$^{2}$The authors are with the Mechanical Engineering Department, Boston University, Boston, MA 02215, USA.
        {\tt\small danyangl@bu.edu, tron@bu.edu}}%
}

\newtheorem{example}{Example}
\usepackage{tikz}

\begin{document}
\def\next{\circ}
\def\always{\square}
\def\event{\lozenge}
\def\relu{\text{ReLU}}

\maketitle
\thispagestyle{empty}
\pagestyle{empty}
\begin{abstract}
Time-series data can represent the behaviors of autonomous systems, such as drones and self-driving cars. The task of binary and multi-class classification for time-series data has become a prominent area of research. Neural networks represent a popular approach to classifying data; However, they lack interpretability, which poses a significant challenge in extracting meaningful information from them. Signal Temporal Logic (STL) is a formalism that describes the properties of timed behaviors. We propose a method that combines all of the above: neural networks that represent STL specifications for multi-class classification of time-series data. We offer two key contributions: 1) We introduce a notion of margin for multi-class classification, and 2) we introduce STL-based attributes for enhancing the interpretability of the results. We evaluate our method on two datasets and compare it with state-of-the-art baselines.
\end{abstract}


\section{INTRODUCTION}
Multi-class classification is an essential problem in Machine Learning (ML). In this paper, we are interested in classifying behaviors from time-series data: given a set of signals sampled over a period of time, we extract features that allow the signal to be assigned to one of the different known categories of interest. Many ML algorithms have been proposed to solve this type of multi-class classification problem; however, most of these algorithms produce models that are hard to interpret by a human, and it is, in general, difficult to identify the attributes (features) of a signal that distinguish one category from the others. Recently, temporal logic tools have attracted lots of attention due to the rich expressiveness and natural interpretability of temporal logic formulae. Signal Temporal Logic (STL) is an expressive formal language to describe the properties of time-series data. The quantitative satisfaction of STL formulae provides a formal way to assess how close a signal is to satisfy a given specification; the specification itself can also be expressed in a language understandable to humans. There have been encouraging results that return STL formulae for classification. However, most of the existing work only considers binary classification tasks~\cite{kong2014temporal,yan2021neural,aasi2022classification}; this is probably due to the fact that the semantics of STL are tightly related to boolean logic, hence offering a natural way to distinguish only between two (positive versus negative) classes. Few studies have been done to introduce the STL technique to a multi-class classifier.

Attribute-based classification is a method intuited by how humans distinguish objects\cite{lampert2009learning}. Humans detect different objects from a high-level description comprising arbitrary semantic attributes. The authors in \cite{lampert2009learning} propose a system that can detect a list of attributes transcending between the classes. The attribute-based classification is statistically efficient since the knowledge about attributes is obtained from different classes. New classes not observed during training can be detected based on their attribute representation without the necessity for an additional training phase, which is also known as zero-shot learning. 
In the domain of autonomous systems, such as robots or drones, their trajectories typically involve the execution of a series of tasks. Tasks often overlap or intersect within a group of autonomous entities to facilitate efficient coordination. The attribute-based classification method categorizes autonomous system trajectories by discerning and learning individual task specifications.
However, most attribute-based classification methods focus on classifying images instead of time-series data\cite{russakovsky2012attribute,li2015multi,lin2019improving}. 

Various techniques have been proposed to solve multi-class classification problems\cite{aly2005survey}. One of the traditional approaches is to convert the multi-class task to multiple binary classification tasks, which can be solved using existing binary classification techniques, such as Support Vector Machine (SVM) \cite{burges1998tutorial}. Others solve the problem by using algorithms that can directly handle multiple classes. For instance, one of the simplest methods is $k$-Nearest Neighbors,  which separates the data into $k$ groups based on the distance to $k$ class of training examples  \cite{bay1998combining}. A more advanced technique that produces more compact models is Decision Trees (DT), which splits the data using sequential decisions across nodes arranged in a tree, and where the leaves of the tree represent class decisions \cite{quinlan2014c4}. Neural networks designate each neuron in the output layer as the identification for one class\cite{abiodun2018state}. However, these techniques produce models that are either hard to interpret or too complex to combine with the idea of attributes.

Efforts have been made to incorporate temporal logic tools to improve the interpretability of the classification models. Existing work includes combining the neural network with weighted signal temporal logic\cite{li2022learning,baharisangari2021weighted,yan2021stone,chen2022interpretable}, extending SVMs to Support Vector Machine-Signal Temporal Logic (SVM-STL)\cite{alsalehi2022learning}, building a decision tree that can be represented as signal temporal logic formulae\cite{aasi2022classification}. These methods, however, focus only on binary classification based on the signs of the quantitative satisfaction of the signal temporal logic formulae. We are aware of two previous works for multi-class classification with DTs STL specifications \cite{linard2022inference, alsalehi2022learning}. Both methods learn the parameters of Parametric Signal Temporal Logic (PSTL). Nevertheless, the DT method is not able to employ attribute learning since a signal ends up at only a single node of DT but may satisfy multiple attributes simultaneously. The efficiency of the decision tree method is restricted by its depth and the sample size of the dataset. Moreover, they only consider the signs (satisfaction or violation) of the STL formulae and discard the quantitative semantics.

\textbf{Contributions:} 
First, we propose a neural network that can be translated to STL formulae for specifying the attributes of signals. Second, we propose a novel margin-based loss function that leverages the quantitative semantics of the STL formulae and explicitly improves the separation between the data. Third, we evaluate our approach on two datasets and highlight the performance of our approach compared to existing methods.

\section{Preliminaries}
\subsection{Error-Correcting Output Codes}\label{sec:ECOC}
The error-correcting output codes (ECOC) technique is a distributed output representation approach in multi-class learning problems\cite{dietterich1994solving}. It has been shown that this output representation is robust with respect to changes in sample size and can reduce overfitting.

The ECOC approach considers the label of each class as a unique codeword, i.e., an arbitrary binary string. The codewords are mapped to a binary coding matrix $E$ with $c$ rows and $n$ columns, where $c$ is the number of classes and $n$ is the number of codeword bits. The length of the codeword, i.e., the number of columns of the coding matrix, must be sufficient to represent $c$ unique labels. The codeword for each class is decided before the training; in general, it is selected to maximize its separation distance from the other codewords. The classifier $f$ is defined by aggregating the output of $n$ binary classifiers $f_1,f_2,\ldots,f_n$, each corresponding to one column of the coding matrix.

The coding matrix $E$ takes the values $\{1,-1\}$ to integrate with the STL inference method. Table \ref{table:ecoc} shows an example of a coding matrix, where $+$ represents $+1$ and $-$ represents $-1$. During the training process, the grounded-truth output of the $i$-th binary classifier $f_i$ is specified by the $i$-th bit of these codewords. For example, $f_2(x)=1$ if $x$ belongs to class $1$ or class $4$, $f_2(x)=-1$ if $x$ belongs to class $2$ or class $3$. The prediction of the classifier $f$ is based on the distance between the codeword obtained by $f_1,\ldots,f_n$ and the codewords in the matrix. In our context, each classifier is correlated to an attribute specified by an STL formula.
\begin{table}[h!]
\centering
\begin{tabular}{ | m{1cm} | m{0.5cm}| m{0.5cm} | m{0.5cm} | m{0.5cm} | }
  \hline
  & $f_1$ & $f_2$ & $f_3$ \\ 
  \hline
  Class 1 & - & +  & -\\ 
  \hline
  Class 2 & - & -  & +\\ 
  \hline
  Class 3 & + & -  & -\\
  \hline
  Class 4 & - & +  & +\\
  \hline
\end{tabular}
\caption{Example of a $3$-bit error-correcting output code for a $4$-class problem.}
\label{table:ecoc}
\end{table}

\subsection{Signal Temporal Logic}
Signal Temporal Logic (STL) is a tool for specifying the spatial-temporal properties of time-series data. We denote a discrete-time, $d$-dimensional, time-series signal $s$ as a finite sequence $s(1),s(2),\ldots,s(l)$, where $s(\tau)\in\real{d}$ is the value of signal $s$ at time $\tau$, $l$ is the length of signal. The syntax of STL formulae is defined recursively as~\cite{maler2004monitoring}:
\begin{equation}\label{eq:stl}
    \varphi::= \pi \mid \neg \varphi \mid \varphi_1\land\varphi_2 \mid \varphi_1\lor\varphi_2 \mid \event_I \varphi \mid \always_I \varphi,
\end{equation}
where $\pi$ is a predicate in the form of $f(s)\sim \mu$, $\mu\in\real{}$, $\sim\in\{\leq,\geq\}$; the time interval $I=[t_1,t_2]$, $t_1,t_2\in\naturals{}$, $t_1\leq t_2$. $\varphi,\varphi_1,\varphi_2$ denote STL formulae. We denote with $\neg,\land,\lor$ the Boolean operators for negation, conjunction, and disjunction, and with $\event,\always$ the temporal operators for eventually operator and always operator.

\begin{definition}[Robustness]
The robustness degree is the quantitative satisfaction of an STL formula given a signal $s$. This quantitative semantics is defined as:
\begin{subequations}\label{stl-semantics}
\begin{align}
    r(s,\pi,\tau) &=f(s(\tau))-\mu \label{stl-semantics:sub1},\\
    r(s,\neg \varphi,\tau) &=-r(s,\varphi,\tau)\label{stl-semantics:sub2},\\
    r(s,\varphi_1\land\varphi_2,\tau) &= \min(r(s,\varphi_1,\tau),r(s,\varphi_2,\tau))\label{stl-semantics:sub3},\\
    r(s,\varphi_1\lor\varphi_2,\tau) &= \max(r(s,\varphi_1,\tau),r(s,\varphi_2,\tau))\label{stl-semantics:sub4},\\
    r(s,\always_I\varphi,\tau) &= \min_{\tau'\in \tau+I}r(s,\varphi,\tau')\label{stl-semantics:sub5},\\
    r(s,\event_I\varphi,\tau) &= \max_{\tau'\in \tau+I}r(s,\varphi,\tau')\label{stl-semantics:sub6},
\end{align}
\end{subequations}
where $\tau+I \in [\tau+t_1,\tau+t_2]$ is a time interval. The real-valued robustness reports the extent to which the signal satisfies or violates an STL specification. Positive robustness ($r(s,\varphi)>0$) indicates that the signal $s$ satisfies the STL formula $\varphi$, denoted as $s\models\varphi$. Negative (or zero) robustness ($r(s,\varphi)\leq 0$) implies the violation of the STL formula $\varphi$ by signal $s$, denoted as $s \not\models \varphi$.
\end{definition}

\subsection{NN-TLI}
The Neural Network (NN) architecture for Temporal Logic Inference (TLI), called NN-TLI~\cite{li2022learning}, is a method for binary classification. The NN-TLI learns an STL formula to classify desired and undesired behaviors as satisfying and violating the formula, respectively. The neural network consists of three layers: a predicate layer, a temporal layer, and a Boolean layer. The predicate layer learns $q_m$ predicates, where $q_m$ is the number of neurons in this layer. The temporal layer learns the time interval corresponding to these $q_m$ predicates. In the Boolean layer, the network formulates the final STL formula using a conjunction-disjunction matrix $M$. The conjunction-disjunction matrix $M$ is a binary matrix defined as:
\begin{equation}\label{eq:c-d matrix}
    M = \bmat{c_{11} &c_{12} &\ldots &c_{1 q_m}\\
    c_{21} &c_{22} &\ldots &c_{2 q_m}\\
    \vdots &\vdots &\ddots &\vdots\\
    c_{p_m 1} &c_{p_m 2} &\ldots &c_{p_m  q_m}\\}\in\{0,1\}^{p_m\times q_m},
\end{equation}
where $q_m$ and $p_m$ are hyperparameters of NN-TLI. The network learns the parameters of matrix $M$; it takes conjunction inside each row, followed by a disjunction between the rows based on the elements of the matrix. The learned STL formula $\phi$ is a \emph{level-1} STL formula in the disjunctive normal form (DNF):
\begin{multline}\label{eq:stl_formula}
    \phi = (\varphi_{11}\land\varphi_{12}\land\ldots\land\varphi_{1q_1})\lor(\varphi_{21}\land\varphi_{22}\land\ldots\land\varphi_{2q_2})\lor\\
    \ldots\lor(\varphi_{p1}\land\varphi_{p2} \land\ldots\land\varphi_{p q_n}),    
\end{multline}
note that $p_m>p$, $q_m\geq\max\{q_1,\ldots,q_n\}$. The conjunction-disjunction matrix only specifies the maximum complexity of the formula, which provides the flexibility to acquire a concise STL formula by learning elements $c_{ij}$'s of matrix $M$. For example, if the predicate layer and temporal layer learn $3$ subformulae $\varphi_1$, $\varphi_2$, $\varphi_3$, and the learned conjunction-disjunction matrix is a $3\times 3$ matrix $M=\bmat{1 &0 & 0\\ 0 &0 &0 \\ 0 &1 &1}$, the final STL formula is $\phi = \varphi_1 \lor (\varphi_2\land\varphi_3)$. For a signal $s$, if $s\models\phi$, it is predicted as the positive label; if $s\not\models\phi$, it is predicted as the negative label. Details of the learning process can be found in \cite{li2022learning}.


\section{Problem Formulation}
We aim to capture the spatial-temporal attributes of signals, learning by STL formulae, and use these attributes for multi-class classification.

In attribute-based classification~\cite{lampert2009learning}, the attribute representation $\mathbf{a}^y = \bmat{a_1^y,\ldots,a_n^y}$ for class $y$ is a fixed-length binary vector. We map the binary attribute representation to an ECOC table and consider the classifier for each attribute as an STL formula. The output of an STL formula is either a positive or non-positive value mapped to an element of $\mathbf{a}^y$. We can then use the prediction method of the ECOC approach to predict the class of a time-series signal.

We define a multi-class classification task with mutually exclusive classes, i.e., where each signal belongs to one and only one class. Consider a dataset with $N$ discrete-time signals $\cS=\{(s^i,\mathbf{a}^i,y^i)\}_{i=1}^N$, where $s^i\in\real{d\times l}$ is the $i$-th signal, $l$ is the length of the signal; $\mathbf{a}^i\in\{-1,+1\}^n$ is the attribute label of the signal; $y^i\in\cC$ is the class of the signal; $\cC=\{1,\ldots,c\}$ is the set of classes, $\abs{\cC}=c$. Our goal is to learn $n$ STL formulae $\phi_1,\phi_2,\ldots,\phi_n$ in the form of \eqref{eq:stl_formula} such that these $n$ STL formulae can be used as the binary classifiers $f's$ explained in Section \ref{sec:ECOC} and decide the class it belongs to. Based on the relation between attributes and classes, we introduce two forms of the problem:
\begin{problem}\label{problem1} (\emph{Attribute-based Classification})
Given a dataset $\cS$, learn $n$ STL formulae $\phi_1,\ldots,\phi_n$, such that:
\begin{equation}
    \forall(s,\mathbf{a},y)\in\cS, (s \models\phi_i, \forall a_i=1) \land (s\not\models\phi_j, \forall a_j=-1).
\end{equation}
\end{problem}

\begin{problem}\label{problem2} (\emph{Class-based Classification})
Given a dataset $\cS$, learn $n$ STL formulae $\phi_1,\ldots,\phi_n$ to construct $\psi_1,\ldots,\psi_c$, such that:
\begin{equation}
    \forall(s,\mathbf{a},y)\in\cS, (s \models\psi_y, y\in \cC) \land (s\not\models\psi_{y'}, \forall y'\in\cC\backslash y),
\end{equation}
where $\psi_i$ is an \emph{class decoder} STL formula defined as:
\begin{equation}
    \psi_i = (\bigwedge\limits_{j,a_j=1} \phi_j)\land (\bigwedge\limits_{k,a_k=-1} \neg \phi_k), \forall i\in\cC.
\end{equation}
Note that if a signal satisfies $\psi_y$, it automatically satisfies the attributes specified by the attribute label $\mathbf{a}^y$, and vice versa; hence, these two problem formulations are semantically the same. However, they lead to different forms for the loss function design.
\end{problem}
\section{Method}
We propose a method for multi-class classification of time-series signals. Our approach is based on three components. First,  we modify the NN-TLI binary classification architecture to a multiclass setting. Second, we introduce a novel notion of margin based on the multi-dimensional embedding produced by the formulae and define a loss function that uses this notion. Finally, we apply \emph{loss-based decoding} based on the error-correcting output codes (ECOC) for the prediction.

\subsection{Multi-class NN-TLI}
Our proposed multi-class NN-TLI architecture is a generalization of the binary NN-TLI architecture. It uses a new conjunction-disjunction matrix $M_m$ for multi-class problems, which we call the \emph{multi-conjunction-disjunction matrix}. For a $c$-class classification problem, we define $M_m$ as:
\begin{equation}
    M_m = \bmat{M_1\\ \vdots\\ M_n}\in\{0,1\}^{n\times p_m \times q_m},
\end{equation}
where $M_i$ is a conjunction-disjunction matrix defined in \eqref{eq:c-d matrix}. Each matrix $M_i$ defines an STL formula $\phi_i$, a Boolean combination of a subset of the subformulae pool. These attributes formulae are learned simultaneously thus can increase training efficiency and accuracy compared to learning $n$ independent neural networks.

\subsection{Multi-class Margin and Margin-based Loss}
For the attribute-based classification (Problem \ref{problem1}), the multi-class NN-TLI can be translated to $n$ STL formulae $\phi_1,\ldots,\phi_n$, and each signal $s$ is associated with an \emph{attribute vector} $\mathbf{r}_a(s)\in\real{n}$ defined as:
\begin{equation}\label{eq:attvector}
\begin{aligned}
    \mathbf{r}_a(s) &=(r(s,\phi_1),r(s,\phi_2),\ldots,r(s,\phi_n))\\
    &=(r_{a1},r_{a2},\ldots,r_{an}).
\end{aligned}
\end{equation}

In the rest of this paper, we denote $r_{ak}$ as the robustness of $\phi_k$ given a sample; $E$ as the ECOC coding matrix; $E(j,\cdot)$ as the $j$-th row of the matrix $E$; $E(\cdot, k)$ as the $k$-th column of the matrix $E$; $E(j,k)$ as the $k$-th bit of the $j$-th row of the matrix $E$. For the class-based classification (Problem \ref{problem2}), we define a \emph{class vector} $\mathbf{r}_y(s)$:
\begin{equation}\label{eq:classvector}
\begin{aligned}
    \mathbf{r}_y(s) &=(r(s,\psi_1),r(s,\psi_2),\ldots,r(s,\psi_c))\\
    &=(r_{y1},r_{y2},\ldots,r_{yc}),
\end{aligned}
\end{equation}
where
\begin{equation}
r(s,\psi_j) = \min_{k=1,\ldots,n} E(j,k)r_{ak}, \forall j\in\cC,
\end{equation}
$r_{yj}$ is the robustness of $\psi_j$ given a sample. The output vector $\mathbf{r}(s)$ of the multi-class NN-TLI is either an attribute vector or a class vector. 

Inspired by the concept of margin in SVM~\cite{wu2005svm}, we introduce a novel notion of margin for multi-class STL inference. In traditional binary SVM classifiers, the margin is defined by the minimum distance between a training sample of either class and the separating hyperplane. The training algorithm then aims to find a hyperplane that maximizes this margin. In our multi-class NN-TLI, instead, we interpret each formula as a hyperplane aligned with the axis. Unlike SVM, the training algorithm does not change the position of this separating hyperplane but changes the position of the output vector $\mathbf{r}$ for each sample. Therefore, we propose a new notion of margin for the multi-class STL classification.

Before providing the formal definition, we first introduce a class table $C\in\{-1,1\}^{c\times c}$, which is a square ECOC matrix generated using one-hot encoding, i.e., the $(i,j)$ entry of matrix $C$ is
\begin{equation}
    C(i,j) = \begin{cases}
    1, & \textrm{if } i=j,\\
    -1, & \textrm{otherwise,}
    \end{cases}
\end{equation}
we then introduce the multi-class STL margin as follows:
\begin{definition}[Multi-class STL margin]
Given $N$ input signals $\{(s^i,\mathbf{a}^i,y^i)\}_{i=1}^N$ and the corresponding output vectors $\{\mathbf{r}(s^i)\}_{i=1}^N$ produced by a multi-class NN-TLI model, the \emph{multi-class STL margin} for attribute-based classification is a vector $\mathbf{m}_a=\bmat{m_{a1},\ldots,m_{an}}$ where
\begin{equation}\label{eq:attmargin}
\begin{gathered}
    m_{ak} = \min_{i} \{\text{ReLU}(E(y^i,k)r_{ak}(s^i))\},\forall k\in\{1,\ldots,n\}\\
    \text{ReLU}(x) = \begin{cases} 
    x, & \textrm{if } x>0,\\
    0, & \textrm{otherwise}.
    \end{cases}
\end{gathered}
\end{equation}
The \emph{multi-class STL margin} for class-based classification is $\mathbf{m}_y=\bmat{m_{y1},\ldots,m_{yc}}$ where
\begin{equation}\label{eq:classmargin}
\begin{aligned}
    &m_{yj} = \min_{i} \{\text{ReLU}(C(y^i,j)r_{yj}(s^i))\},\forall j\in\{1,\ldots,c\}.
\end{aligned}
\end{equation}
\end{definition}

\begin{figure}[ht]
  \centering
  \subfloat[The three-dimensional output space]{\includegraphics[scale=0.28]{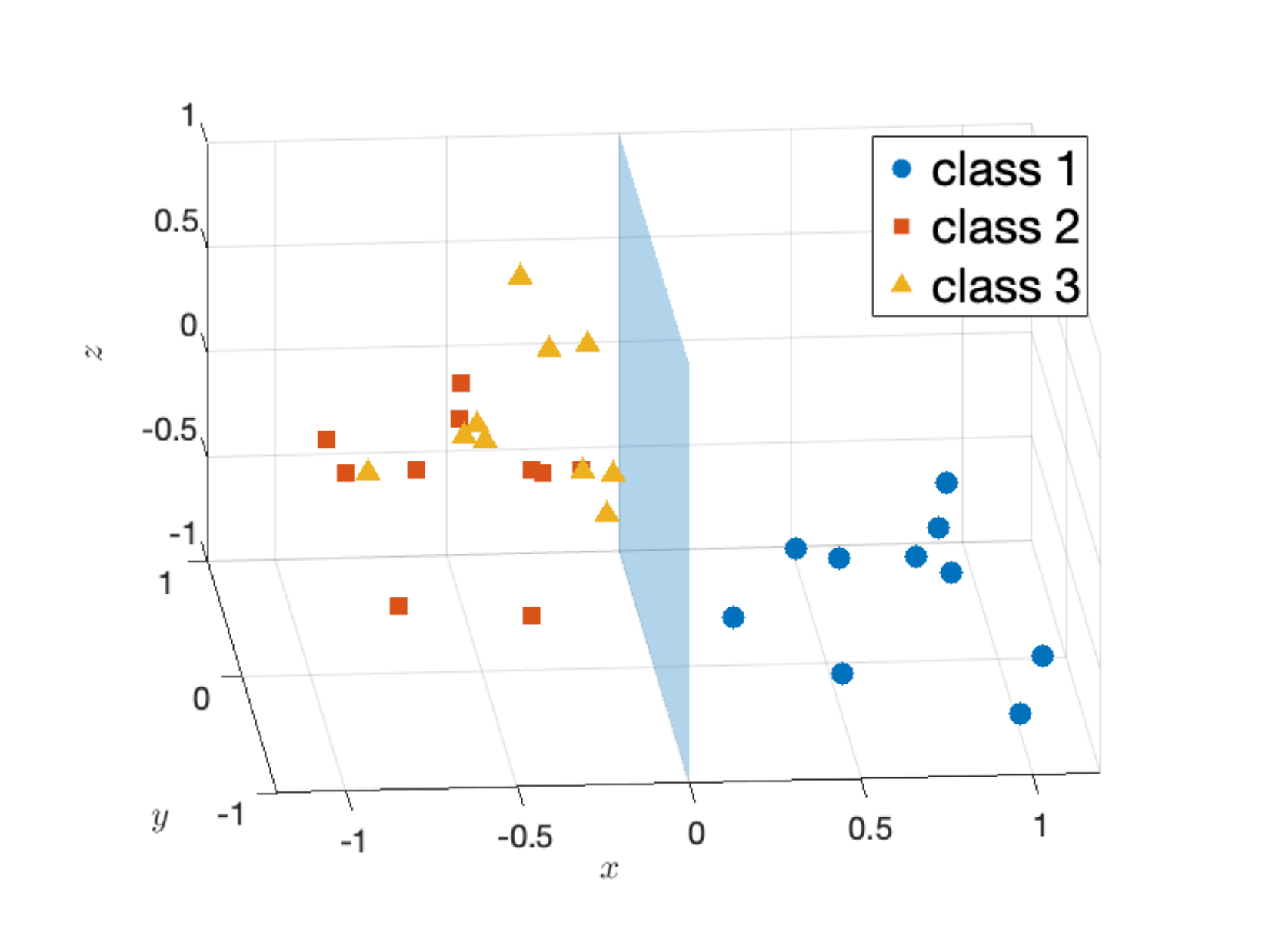}\label{fig:s1}}
  \hfill
  \subfloat[The projection onto one-dimensional subspace]{\hspace*{-4mm}\includegraphics[scale=0.22]{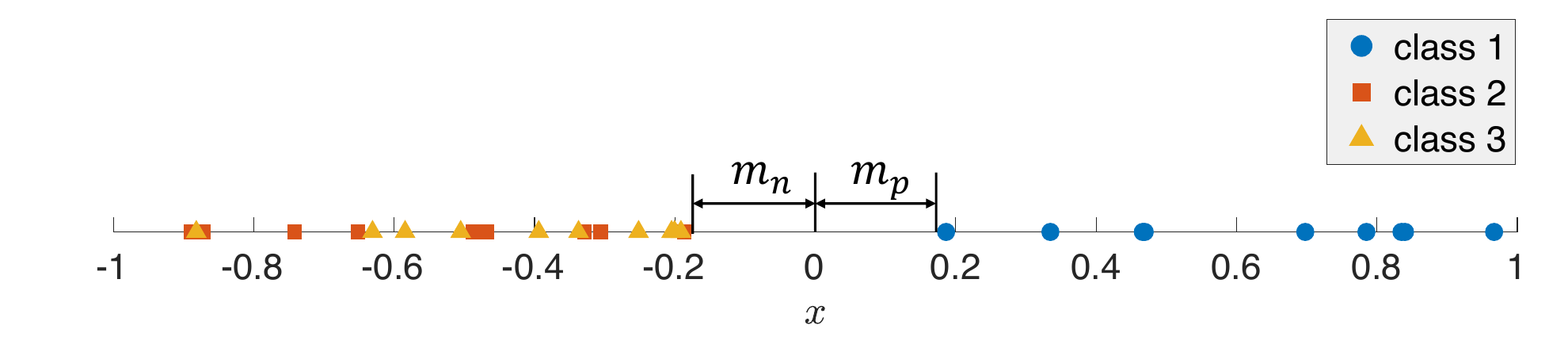}\label{fig:s2}}
  \caption{Example of a three-dimensional output space.}
  \label{figure:margin}
\end{figure}

The margin of each STL formula can be interpreted as a distance from the correctly classified points to the origin in a subspace projected onto the coordinate axes. The margin can be considered as a quantitative measurement of the separation of signals. We can use the margin to find more expressive formulae to separate the data.

\begin{example}
Consider a three-class classification task with a class set $\{1,2,3\}$ and a class-based classification. The output space created by multi-class NN-TLI is $\real{3}$. The network maps the training samples to three-dimensional points, as shown in Figure \ref{fig:s1}. We can project the points to one of the axes, e.g., the $x$-axis subspace shown in Figure \ref{fig:s2}. From the problem statement, the points with positive $x$-coordinate satisfy $\psi_1$, and those with negative $x$-coordinate violate $\psi_1$. The margin of $\psi_1$ is $m_{y1}=\min\{m_p,m_n\}$.
\end{example}

We design the loss function to satisfy two requirements: (1) The loss is small if the inferred formula is satisfied by positive data and violated by negative data. (2) The loss is small if the inferred formula can achieve a large margin. The loss function for the attribute-based classification problem for dataset $\cS$ is defined as
\begin{multline}
    L_a(\mathbf{r}_a,E) = \sum_{k=1}^{n} \sum_{i=1}^N \Big(\text{ReLU}\big(m_{ak} - E(y^i,k) r_{ak}(s^i)\big)\\
    -\delta m_{ak}\Big),
\end{multline}
while the loss function for the class-based classification problem is
\begin{multline}
    L_y(\mathbf{r}_y,C) = \sum_{j=1}^{c} \sum_{i=1}^N \Big(\text{ReLU}\big(m_{yj} - C(y^i,j) r_{yj}(s^i)\big)\\
    -\delta m_{yj}\Big),
\end{multline}
where $\delta>0$ is a tuning parameter used to control the compromise between maximizing the margin and classifying more data correctly. If $\delta$ is large, the learned STL formulae tend to maximize the margin, but it may cause more samples to be misclassified. If $\delta$ is small, a formula might exist to better separate the positive and negative data.

\subsection{Loss-based Decoding}\label{sec:loss}
There are many ways to predict the class of a new sample in multi-class problems. A widely used method is classifying the example as the class whose label has the "smallest" distance to the prediction.

For brevity of annotation in this section, we use $\mathbf{r}(s)$ to denote either $\mathbf{r}_a(s)$ or $\mathbf{r}_y(s)$ and $T$ to denote either $E$ or $C$, respectively. The signal $s$ is classified as class $\hat{y}$ according to:
\begin{equation}
    \hat{y} = \argmin_j d(\mathbf{r}(s),T(j,\cdot)),
\end{equation}
where $d(\cdot)$ is a distance measurement. In this way, the method for measuring the distance between the prediction and the label will significantly affect the prediction.

\subsubsection{Hamming distance}
The commonly used Hamming distance $d_H$ is defined as:
\begin{equation}
    d_H(\mathbf{r}(s),T(j,\cdot)) = \sum_k(1-\text{sign}(T(j,k)r_k(s))),
\end{equation}
where $r_k(s)$ is the $k$-th element of $\mathbf{r}(s)$. The Hamming distance calculates the number of bits of $T(j,\cdot)$ whose sign differs from those of $\mathbf{r}(s)$. This method only considers the sign of elements of $\mathbf{r}(s)$ and discards the robustness information provided by STL formulae.

\subsubsection{Loss-based distance}
We use another way of measuring the distance called loss-based distance\cite{Allwein2000ReducingMT},
\begin{equation}\label{eq:lossdistance}
    d_L(\mathbf{r}(s),T(j,\cdot)) = \sum_k \text{ReLU}(-T(j,k) r_k(s)).
\end{equation}
Since the margin is introduced only to improve the classifier during training, the loss-based distance is the loss function while removing the margin. As a result, the notion of the margin is not used in the prediction.

\begin{algorithm}[ht]
  \begin{algorithmic}[1]
    \Require Time-series dataset $\cS$, ECOC coding matrix $E$, number of iterations $K$.
    \Ensure $n$ STL formulae $\phi_1,\ldots,\phi_{n}$.
    \State Initialize the parameters of multi-class NN-TLI.
    \For{$i\in{1,\ldots,K}$}
    \State Select a batch data from $\cS$.
    \State Compute the robustness of $\phi_1,\ldots,\phi_{n}$.
    \State Compute the loss.
    \State Perform backpropagation on the loss.
    \State Update the parameters of the multi-class NN-TLI.
    \EndFor
    \State \Return $\phi_1,\ldots,\phi_{n}$.
  \end{algorithmic}
  \caption{Multi-class NN-TLI Learning Algorithm.}
  \label{algo:network}
\end{algorithm}

\section{Case study}
In this section, we first implement our approach to classify and interpret multiple behaviors in a naval scenario using STL formulae and show how we can learn interpretable attributes of each class using the ECOC coding matrix. We also compare with state-of-the-art methods on a synthetic dataset and demonstrate the benefits of our method. For all the case studies in this section, we test both the attribute-based classification (Problem \ref{problem1}) and the class-based classification methods (Problem \ref{problem2}). The mean and variance of training time and misclassification rate are similar for both methods, which implies that there is no fundamental difference between these two methods. Due to the space limitation, we only report results from the attribute-based classification in this section.

\subsection{Naval Scenario Dataset}
\begin{figure}[ht]
    \centering
    \includegraphics[scale=0.35]{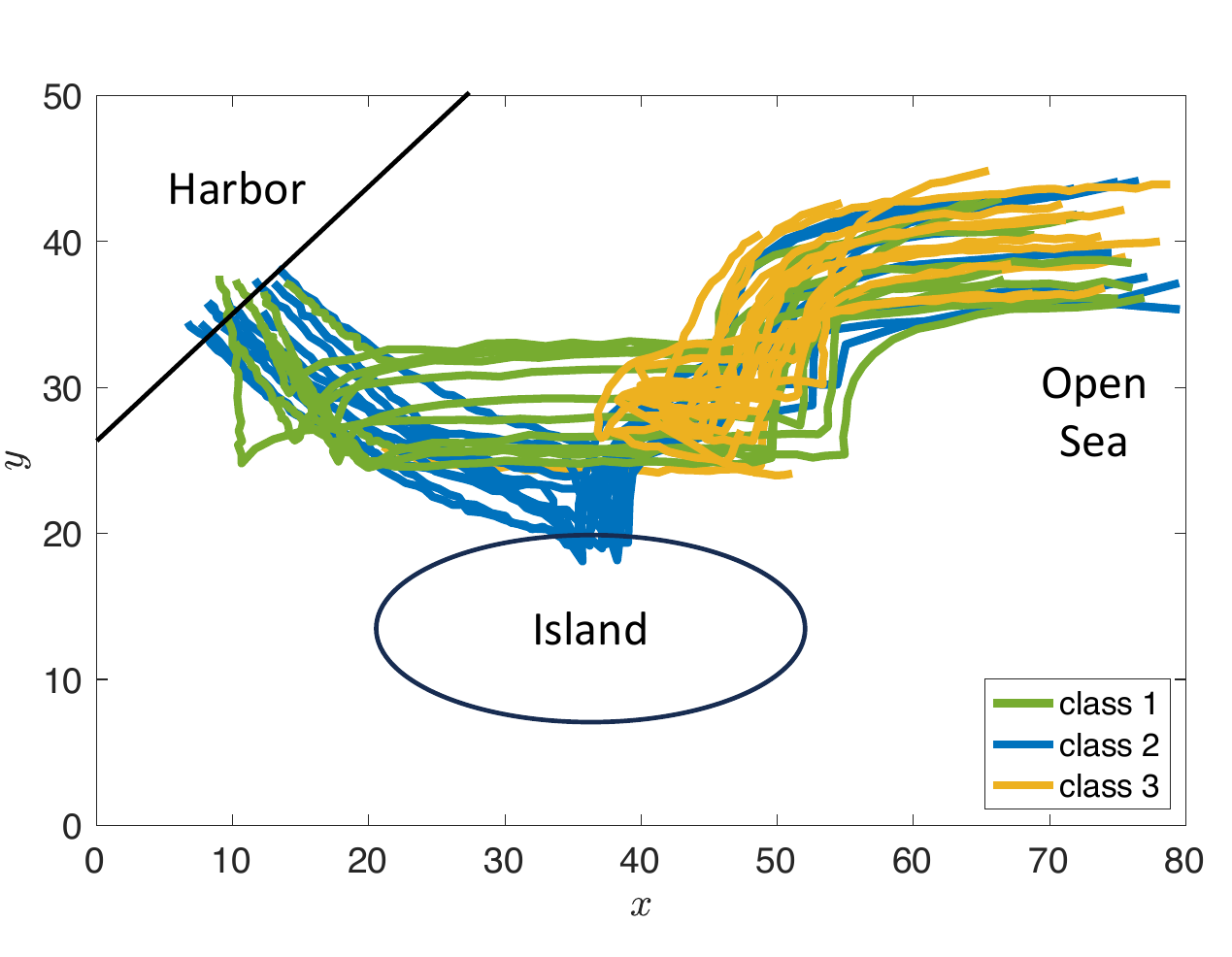}
    \caption{Trajectories in the naval scenario. The green trajectories are labeled as class $1$. The blue trajectories are labeled as class $2$. The yellow trajectories are labeled as class $3$.}
    \label{figure:naval}
\end{figure}

We use a dataset from \cite{aasi2022classification} related to a naval surveillance scenario. The dataset consists of three kinds of vessel behaviors. The trajectories are two-dimensional and have $60$ time steps in total. As shown in Fig \ref{figure:naval}, the trajectories are divided into $3$ classes described as class 1: ``arrive harbor while not reaching island"; class $2$: ``arrive harbor while passing through island"; class $3$: ``return to the open sea without reaching harbor or island." Class $1$ has $1000$ trajectories, class $2$ and class $3$ have $500$ trajectories each.

In this experiment, we use two kinds of coding matrices. The first kind of matrix (Table \ref{table:ECOC naval class}) is coded based on the class. In this case, traditional class learning is applied, but not attribute learning; each binary classifier $f$ is used to classify one class. We do not learn the attributes but only the property of each class. The second kind of matrix (Table \ref{table:ECOC naval attribute1},\ref{table:ECOC naval attribute2},\ref{table:ECOC naval attribute3}) is coded based on the attributes. Each binary classifier $f$ is used to classify one attribute. The trajectories are classified based on whether they contain certain attributes. For example, in $E_{a1}$, class $1$ has attribute $1$ but not attribute $2$; class $2$ has both attribute $1$ and $2$; class $3$ has neither attribute $1$ or $2$. According to \cite{dietterich1994solving}, there are at most $2^{c-1}-1$ usable columns for a $c$-class problem after excluding the complementary, all-zeros or all-ones columns. We used different attribute-based coding matrices $E_{a1}$, $E_{a2}$, and $E_{a3}$ to show that we can achieve semantically equivalent results regardless of the design of the coding matrix.

We test our algorithm on a 5-fold cross-validation and report the average training time and average misclassification rate $MCR$. The misclassification rate $MCR$ is defined as:
\begin{equation*}
    MCR = \frac{\lvert\{\{s,\mathbf{a},y\}\mid (s \models \phi_i, a_i=-1 \lor s\not\models \phi_i, a_i=1)\}\rvert}{N},
\end{equation*}
where $N$ is the number of samples in the dataset. The initialization of the multi-class NN-TLI, the batch size, and the number of training epochs are consistent for all cases. The results are shown in Table \ref{table:naval class result}-\ref{table:naval attribute3 result}. Using the attribute-based coding matrix, we can reach zero misclassification rate with less training time since there are fewer binary classifiers to learn compared to the class-based coding matrix. The attribute learning can also provide more concise STL formulae to interpret the features that are easy for humans to understand. 

\begin{table}[ht]
\hfill
\centering
    \subfloat[Class-based coding matrix $E_c$]{
    \label{table:ECOC naval class}
    \begin{tabular}{ | m{1.0cm} | m{0.4cm}| m{0.4cm} | m{0.4cm} | }
      \hline
      &$f_1$ & $f_2$ & $f_3$ \\ 
      \hline
      Class $1$ &+ &- &-\\ 
      \hline
      Class $2$ &- &+ & -\\ 
      \hline
      Class $3$ &- &- &+\\
      \hline
    \end{tabular}
    }
    \quad
    \subfloat[The training time and accuracy(class-based)]{
        \label{table:naval ta}
        \begin{tabular}{ | m{1.0cm}| m{1.0cm} |}
          \hline
          time(s) &MCR\\ 
          \hline
          23.69 & 0.00\\
          \hline
        \end{tabular}
    }
    \hfill
    \subfloat[The learned STL formulae for each class]{
        \label{table:naval STL}
        \centering
        \begin{tabular}{ c  m{4.0cm} m{2.0cm} }
        \toprule
        Class &STL formulae &Description\\
        \midrule
        Class 1 &$\event_{[2,60]}(x<22.5) \land \always_{[7,45]}(y>23.4)$ &reach harbor and not reach island\\
        \midrule
        Class 2 &$\event_{[10,49]}(y<20.0)$ &reach island\\
        \midrule
        Class 3 &$\always_{[7,43]}(x>33.8)$ &stay in open sea\\
        \bottomrule
        \end{tabular}
     }
\caption{Cross-validation results learned from class-based coding matrix $E_c$.}
\label{table:naval class result}
\end{table}

\begin{table}[ht]
\hfill
    \centering
    \subfloat[Attribute-based coding matrix $E_{a1}$]{
    \label{table:ECOC naval attribute1}
    \begin{tabular}{ | m{1.0cm} | m{0.9cm}| m{0.9cm} | }
      \hline
      &$f_1$ & $f_2$ \\ 
      \hline
      Class $1$ &+ &- \\ 
      \hline
      Class $2$ &+ &+ \\ 
      \hline
      Class $3$ &- &- \\
      \hline
    \end{tabular}
    }
    \quad
    \subfloat[The training time and accuracy ($E_{a1}$)]{
        \label{table:naval attribute1 ta}
        \begin{tabular}{ | m{1.0cm}| m{1.0cm} |}
          \hline
          time(s) &MCR\\ 
          \hline
          16.14 & 0.00\\ 
          \hline
        \end{tabular}
    }
    \hfill
    \subfloat[The learned STL formulae for each attribute in $E_{a1}$]{
        \label{table:naval attribute1 STL}
        \begin{tabular}{c c c}
        \toprule
        Attribute &STL formulae &Description\\
        \midrule
        Attribute 1 &$\event_{[7,60]}(x<22.1)$ &reach harbor\\
        \midrule
        Attribute 2 &$\event_{[0,48]}(y<19.8)$ &reach island\\
        \bottomrule
        \end{tabular}
     }
\caption{Cross-validation results learned from attribute-based coding matrix $E_{a1}$.}
\label{table:naval attribute1 result}
\end{table}

\begin{table}[ht]
\hfill
    \centering
    \subfloat[Attribute-based coding matrix $E_{a2}$]{
    \label{table:ECOC naval attribute2}
    \begin{tabular}{ | m{1.0cm} | m{0.9cm}| m{0.9cm} | }
      \hline
      &$f_1$ & $f_2$ \\ 
      \hline
      Class $1$ &+ &+ \\ 
      \hline
      Class $2$ &+ &- \\ 
      \hline
      Class $3$ &- &- \\
      \hline
    \end{tabular}
    }
    \quad
    \subfloat[The training time and accuracy ($E_{a2}$)]{
        \label{table:naval attribute2 ta}
        \begin{tabular}{ | m{1.0cm}| m{1.0cm} |}
          \hline
          time(s) &MCR\\ 
          \hline
          15.85 & 0.00\\ 
          \hline
        \end{tabular}
    }
    \hfill
    \subfloat[The learned STL formulae for each attribute in $E_{a2}$]{
        \label{table:naval attribute2 STL}
        \begin{tabular}{c m{3.0cm} m{2.0cm}}
        \toprule
        Attribute &STL formulae &Description\\
        \midrule
        Attribute 1 &$\event_{[6,60]}(x<23.1)$ &reach harbor\\
        \midrule
        Attribute 2 &$\event_{[4,60]}(x<22.3)\land\always_{[8,44]}(y>23.4)$ &reach harbor and not reach island\\
        \bottomrule
        \end{tabular}
     }
\caption{Cross-validation results learned from attribute-based coding matrix $E_{a2}$.}
\label{table:naval attribute2 result}
\end{table}

\begin{table}[ht]
\hfill
\centering
    \subfloat[Attribute-based coding matrix $E_{a3}$]{
    \label{table:ECOC naval attribute3}
    \begin{tabular}{ | m{1.0cm} | m{0.9cm}| m{0.9cm} | }
      \hline
      &$f_1$ & $f_2$ \\ 
      \hline
      Class $1$ &+ &+ \\ 
      \hline
      Class $2$ &- &- \\ 
      \hline
      Class $3$ &+ &- \\
      \hline
    \end{tabular}
    }
    \quad
    \subfloat[The training time and accuracy ($E_{a3}$)]{
        \label{table:naval attribute3 ta}
        \begin{tabular}{ | m{1.0cm}| m{1.0cm} |}
          \hline
          time(s) &MCR\\ 
          \hline
          15.86 & 0.00\\ 
          \hline
        \end{tabular}
    }
    \hfill
    \subfloat[The learned STL formulae for each attribute in $E_{a3}$]{
        \label{table:naval attribute3 STL}
        \begin{tabular}{c m{3.0cm} m{2.0cm}}
        \toprule
        Attribute &STL formulae &Description\\
        \midrule
        Attribute 1 &$\always_{[9,55]}(y>22.4)$ &not reach island\\
        \midrule
        Attribute 2 &$\always_{[4,45]}(y>23.3)\land\event_{[0,60]}(x<24.0)$ & not reach island and reach harbor\\
        \bottomrule
        \end{tabular}
     }
     \caption{Cross-validation results learned from attribute-based coding matrix $E_{a3}$.}
     \label{table:naval attribute3 result}
\end{table}

\subsection{Synthetic Dataset}
\begin{figure}[ht]
    \centering
    \hspace*{-8mm}
    \includegraphics[scale=0.42]{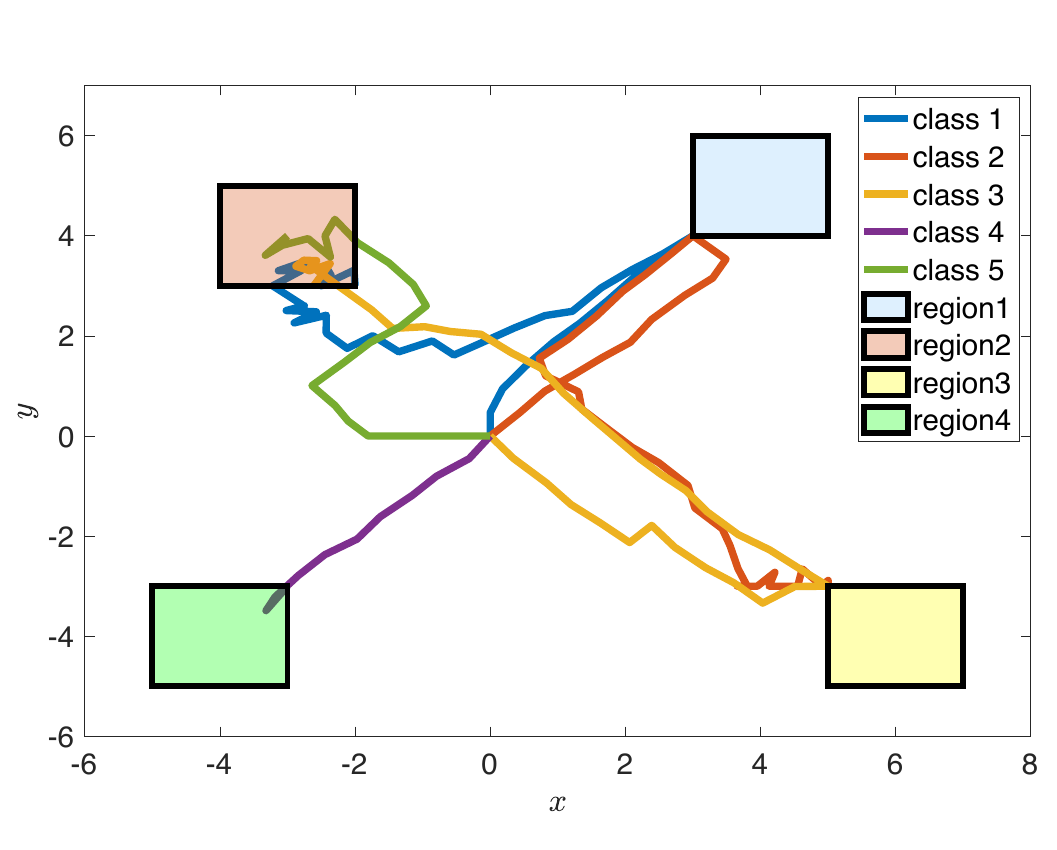}
    \caption{Examples of trajectories in the synthetic dataset. Region $1$, $2$, $3$, $4$ corresponds to the region specified by the predicate of $\phi_1$, $\phi_2$, $\phi_3$, $\phi_4$, respectively.}
    \label{figure:synthetic}
\end{figure}

In this example, we implement experiments on a synthetic dataset and compare it with the state-of-the-art methods~\cite{multiDT,gers2000learning}. The problem formulation in our paper is different from that in \cite{multiDT}; the ``multi-label" of a signal in \cite{multiDT} is equivalent to attributes in our paper, so we generate a new dataset for better comparison. The trajectories in the synthetic dataset are 2-dimensional time-series data generated by four different STL specifications using a MILP approach. Since the STL-difference method in \cite{multiDT} can only deal with STL formulae with the predicate in the form:
\begin{equation}
    p= (\alpha\leq x\leq \beta) \land (\gamma \leq y\leq \lambda),\\
\end{equation}
the STL specifications to generate the dataset are:
\begin{equation}
    \begin{aligned}
        \phi_1 &= \event_{[0,10]}(3\leq x\leq 5\land 4\leq y\leq 6),\\
        \phi_2 &= \always_{[30,40]}(-4\leq x\leq -2\land 3\leq y\leq 5),\\
        \phi_3 &= \event_{[10,40]}(5\leq x\leq 7\land -5\leq y\leq -3),\\
        \phi_4 &= \event_{[10,40]}(-5\leq x\leq -3\land -5\leq y\leq -3).
    \end{aligned}
\end{equation}
Figure \ref{figure:synthetic} shows some example trajectories in the synthetic dataset. The signal is required to arrive at some regions within the desired time interval specified by the STL specifications corresponding to the class. The synthetic dataset contains 5 classes labeled as $c_1$, $c_2$, $c_3$, $c_4$, $c_5$ with trajectories individually satisfy:
\begin{equation}
    \begin{aligned}
        c_1 &= \phi_1\land\phi_2\land\neg\phi_3\land\neg\phi_4,\\
        c_2 &= \phi_1\land\phi_3\land\neg\phi_2\land\neg\phi_4,\\
        c_3 &= \phi_2\land\phi_3\land\neg\phi_1\land\neg\phi_4,\\
        c_4 &= \phi_4\land\neg\phi_1\land\neg\phi_2\land\neg\phi_3,\\
        c_5 &= \phi_2\land\neg\phi_1\land\neg\phi_3\land\neg\phi_4.
    \end{aligned}
\end{equation}

\begin{table*}[ht]
\centering
\hfill
\begin{tabular}
{||m{0.35cm}|m{0.7cm}|m{0.4cm}|m{0.3cm}|m{0.7cm}|m{0.4cm}|m{0.3cm}|m{0.7cm}|m{0.4cm}|m{0.3cm}|m{0.7cm}|m{0.4cm}|m{0.3cm}|m{0.7cm}|m{0.4cm}|m{0.3cm}|m{0.7cm}|m{0.4cm}|m{0.3cm}||}
\hline
  &\multicolumn{6}{|c|}{sample size = $500$} &\multicolumn{6}{|c|}{sample size = $2500$} &\multicolumn{6}{|c|}{sample size = $5000$}\\ 
  \hline
  &\multicolumn{3}{|c|}{$MCR$} &\multicolumn{3}{|c|}{time(s)} &\multicolumn{3}{|c|}{$MCR$} &\multicolumn{3}{|c|}{time(s)} &\multicolumn{3}{|c|}{$MCR$} &\multicolumn{3}{|c|}{time(s)} \\ 
  \hline
  fold &$mSTL$ &$dt\Delta$ &$rnn$ &$mSTL$ &$dt\Delta$ &$rnn$ &$mSTL$ &$dt\Delta$ &$rnn$ &$mSTL$ &$dt\Delta$ &$rnn$ &$mSTL$ &$dt\Delta$ &$rnn$ &$mSTL$ &$dt\Delta$ &$rnn$\\
  \hline
  1 &0.00 &0.00 &0.00 &64 &324 &67    &0.00 &0.00 &0.00 &63 &1689 &69    &0.00 &0.00 &0.00 &65 &3236 &67\\ 
  \hline
  2 &0.00 &0.00 &0.00 &63 &299 &68    &0.00 &0.00 &0.00 &62 &1431 &68    &0.00 &0.00 &0.00 &64 &3220 &67\\
  \hline
  3 &0.00 &0.00 &0.00 &65 &427 &68    &0.00 &0.00 &0.00 &62 &1786 &73    &0.00 &0.00 &0.00 &64 &3360 &66\\
  \hline
  4 &0.00 &0.00 &0.00 &63 &395 &66    &0.00 &0.00 &0.00 &63 &1594 &68    &0.00 &0.00 &0.00 &63 &3821 &67\\
  \hline
  5 &0.00 &0.00 &0.00 &63 &359 &67    &0.00 &0.00 &0.00 &63 &1931 &65   &0.00 &0.00 &0.00 &63 &3353 &67\\
  \hline
   &0.00 &0.00 &0.00 &64 &361 &67    &0.00 &0.00 &0.00 &63 &1686 &69    &0.00 &0.00 &0.00 &64 &3398 &67\\
  \hline
\end{tabular}
\caption{Cross validation result of multi-class NN-TLI ($mSTL$), decision tree-based algorithm ($dt\Delta$) and recurrent neural network ($rnn$) on datasets with different sample size.}
\label{table:cross-val}
\end{table*}

We conduct experiments on the datasets with sample sizes of $500$, $2500,$ and $5000$; these three datasets have $100$, $500,$ and $1000$ samples for each class, respectively. We also compare the performance with the decision-tree-based algorithm \cite{multiDT} and a recurrent neural network \cite{gers2000learning} on these datasets. The decision-tree-based method ($dt\Delta$) uses a decision tree with a maximum depth of 5. The recurrent neural network (denoted as $rnn$ in Table \ref{table:cross-val}) has $4$ hidden layers, with the first layer as an LSTM layer.

\begin{figure}[ht]
    \centering
    \includegraphics[scale=0.45]{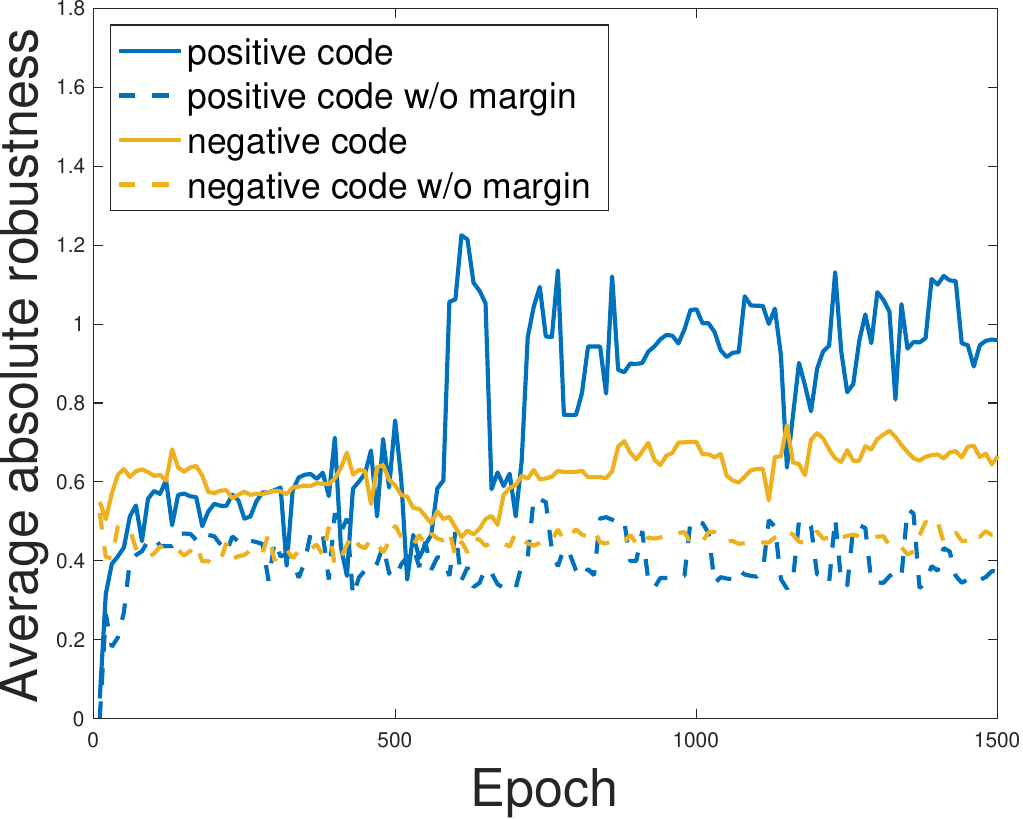}
    \caption{The average absolute robustness value versus the epoch for positive and negative code, with and without margin, respectively.}
    \label{figure:robustness}
\end{figure}

Table~\ref{table:cross-val} shows the results of these three methods on a 5-fold cross-validation. The initialization of the multi-class NN-TLI is consistent for each fold. The batch size and the number of training epochs are the same for neural networks, i.e., multi-class NN-TLI and $rnn$, across all folds. The ECOC matrix we use is the class-based coding matrix since $dt\Delta$ algorithm cannot implement the attribute learning. The results show that the multi-class NN-TLI, called $mSTL$ for short reference, can learn the STL formulae describing the properties of each class and achieve a low misclassification rate with high efficiency. The $mSTL$ and $rnn$ methods can achieve much higher efficiency than the $dt\Delta$, while $rnn$ cannot provide the characteristics of trajectories of each class. The decision tree-based algorithm $dt\Delta$ is able to learn the STL formulae. However, the training time of $dt\Delta$ increases with the sample size of the dataset. The reason for that is the computational bottleneck of the decision tree method, as it iterates all the input instances to find the optimal split at a given node. This step is repeated for each node of the tree \cite{dtbottlenck}. Thus, the computational time grows with the total number of instances. However, the neural network method iterates over the selected batch data at each epoch, which is small relative to the total number of instances.

\begin{table}[ht]
\centering
    \subfloat[The coding matrix $E_{obs}$]{
    \label{table:zeroobserve}
    \begin{tabular}{ | m{0.7cm} | m{0.5cm}| m{0.5cm} | m{0.5cm} | m{0.5cm} | }
      \hline
      &$f_1$ & $f_2$ & $f_3$ &$f_4$ \\ 
      \hline
      $c_1$ &+ &+ &- &-\\ 
      \hline
      $c_2$ &+ &- &+ &-\\ 
      \hline
      $c_3$ &- &+ &+ &-\\
      \hline
      $c_4$ &- &- &- &+\\
      \hline
    \end{tabular}
    }
    \hfill
    \subfloat[The coding matrix $E_{pred}$]{
    \label{table:zeropredict}
    \begin{tabular}{ | m{0.7cm} | m{0.5cm}| m{0.5cm} | m{0.5cm} | m{0.5cm} | }
      \hline
      &$f_1$ & $f_2$ & $f_3$ &$f_4$ \\ 
      \hline
      $c_5$ &- &+ &- &-\\ 
      \hline
    \end{tabular}
    }
\caption{The ECOC coding matrix for zero-shot learning.}
\label{table:attributezero}
\end{table}

To investigate the effectiveness of the margin in the loss function during the training process, we recorded the average absolute robustness value of the correctly classified trajectories shown in Fig~\ref{figure:robustness}. For comparison, we also trained a multi-class NN-TLI without margin in the loss function and recorded the average absolute robustness, also shown in Fig~\ref{figure:robustness}. It can be observed that the introduction of the margin to the loss can increase the absolute value of the average robustness. In other words, the margin can improve the separation between the trajectories during the training process and achieve better performance.

We also implement a zero-shot learning~\cite{xian2018zero} to demonstrate the effectiveness of attribute-based recognition in our multi-class NN-TLI framework. We extract attributes from observed samples and employ them to predict the class of non-observed data through attribute composition. We generate $5$ classes of samples based on $4$ types of attributes, with their labels denoting possession of attributes, as shown in Table~\ref{table:attributezero}. The multi-class NN-TLI is trained over observed classes $c_1$, $c_2$, $c_3$, $c_4$ using the coding matrix $E_{obs}$ in Table \ref{table:zeroobserve} and learn $4$ binary classifier $f_1$, $f_2$, $f_3$, $f_4$ correlating to $4$ kinds of attributes. These four attributes are interpreted by four STL formulae:
\begin{equation}
    \begin{aligned}
        \psi_1 &=\event_{[0,3]}(-0.1<x<2.4\land 0.3<y<2.4),\\
        \psi_2 &=\always_{[26,27]}(-3.9<x<0.6\land 1.0<y<4.1),\\
        \psi_3 &=\event_{[23,28]}(-0.4<x<5.0\land -5.3<y<2.6),\\
        \psi_4 &=\event_{[1,20]}(-4.1<x<-0.2\land -3.8<y<-0.5),\\
    \end{aligned}
\end{equation}
we then use these attributes to predict the unobserved class $c_5$ labeled as $E_{pred}$ in Table \ref{table:zeropredict}. We can achieve $0.00$ $MCR$ for classifying the trajectories in $c_5$. The multi-class NN-TLI can capture the attributes of the signals and interpret the characteristics through the learned STL formulae.

\section{Conclusions and Future Work}
We propose a multi-class NN-TLI neural network designed for multi-class classification tasks on time-series data. The neural network can learn STL formulae to interpret signal attributes and classify signals based on their attribute representation. Our method is efficient, offering analysis from a general perspective and providing concise descriptions of the data. In the future, we will focus on determining the importance of attributes by learning the weight of each attribute.





\bibliographystyle{IEEEtran}
\bibliography{Multiclass}

\end{document}

%% file: preamble/common.tex
\input{preamble/commonNoTikz}
\input{preamble/graphicsTikz}
\input{preamble/robotics}


%% file: preamble/commonNoTikz.tex
\input{preamble/fixes}
\input{preamble/graphics}
\input{preamble/pagination}
\input{preamble/utilities}
\input{preamble/math}
\input{preamble/operators}

\input{preamble/notation}

\input{preamble/units}

\input{preamble/markupAndCommenting}


%% file: preamble/fixes.tex
\usepackage{fix-cm}
\usepackage{etex}

\usepackage{dblfloatfix}

\usepackage{nag}


%% file: preamble/graphics.tex
\makeatletter
\@ifpackageloaded{xcolor}{}{%
\usepackage[table,x11names,dvipsnames,svgnames]{xcolor}%
}
\makeatother

\usepackage{colortbl}

\usepackage{graphicx}
\usepackage{wrapfig}

\input{preamble/graphicsColors}


%% file: preamble/graphicsColors.tex
\definecolorset{RGB}{lyft}{}{Red,194,39,36;Sunset,202,53,33;Orange,205,68,20;Amber,200,117,42;Yellow,242,169,52;Citron,186,188,44;Lime,112,159,33;Green,56,139,31;Mint,45,118,56;Teal,52,133,135;Cyan,60,132,202;Blue,55,94,248;Indigo,64,13,247;Purple,115,42,248;Pink,176,25,145;Rose,176,32,75}

%% file: preamble/pagination.tex
\usepackage{cite}

\usepackage{microtype}

\usepackage{array}
\usepackage{multirow}
\usepackage{booktabs}
\usepackage{makecell} 

\ifcsname labelindent\endcsname
\let\labelindent\relax
\fi
\usepackage[inline]{enumitem}

\usepackage{subfig}

\newenvironment{lenumerate}[2][]
{\begin{enumerate}[label=(#2\arabic*),leftmargin=0.2in,itemindent=0.15in,#1]}
{\end{enumerate}}

\setlist*[enumerate,1]{label={\itshape\arabic*)}}

\makeatletter
\newcommand{\paragraphswithstop}{%
\let\copyparagraph\paragraph%
\renewcommand\paragraph[1]{\copyparagraph{##1.}}%
}
\makeatother

\usepackage[framemethod=tikz]{mdframed}


%% file: preamble/utilities.tex
\usepackage{suffix}

\usepackage{environ}

\makeatletter
\newsavebox{\boxifnotempty}
\newcommand{\displayifnotempty}[3]{\sbox\boxifnotempty{#2}\setbox0=\hbox{\usebox{\boxifnotempty}\unskip}%
\ifdim\wd0=0pt
\else
 #1\usebox{\boxifnotempty}#3%
\fi%
}
\newcommand{\ifempty}[2]{\setbox0=\hbox{#1\unskip}%
\ifdim\wd0=0pt%
 #2%
\fi%
}
\newcommand{\ifnotempty}[2]{\setbox0=\hbox{#1\unskip}%
\ifdim\wd0>0pt%
 #2%
\fi%
}
\makeatother

\usepackage{algpseudocode}
\usepackage{algorithm}

\usepackage{scrlfile}

\makeatletter
\newcommand*\newstoreddef[1]{
  \BeforeClosingMainAux{%
    \immediate\write\@auxout{%
      \string\restoredef{#1}{\csname #1\endcsname}%
    }%
  }%
}
\newcommand*{\restoredef}[2]{
  \expandafter\gdef\csname stored@#1\endcsname{#2}%
}
\newcommand*{\storeddef}[1]{
  \@ifundefined{stored@#1}{0}{\csname stored@#1\endcsname}%
}
\makeatother

\usepackage{pageslts}
\NewEnviron{tee}{\BODY\typeout{Marker Tee [start] ^^J \BODY ^^JMaker Tee [end]}}


%% file: preamble/operators.tex
\newcommand{\real}[1]{\mathbb{R}^{#1}{}}

\newcommand{\naturals}[1]{\mathbb{N}^{#1}{}}

\newcommand{\bmat}[1]{\begin{bmatrix}#1\end{bmatrix}}

\DeclarePairedDelimiter{\abs}{\lvert}{\rvert}

\DeclareMathOperator*{\argmin}{\arg\!\min}



%% file: preamble/notation.tex
\providecommand{\cC}{\mathcal{C}}

\providecommand{\cS}{\mathcal{S}}


%% file: preamble/units.tex
\usepackage{units}


%% file: preamble/markupAndCommenting.tex
\newcommand{\newcolorlabel}[2]{%
  \expandafter\newcommand\csname #1\endcsname[1]{%
    \colorbox{#2}{\color{white}\textsf{\textbf{##1}}}}%
}

\newcommand{\newcommenter}[2]{%
  \expandafter\newcommand\csname #1\endcsname[1]{%
    \fcolorbox{#2}{#2}{\color{white}\textsf{\textbf{#1}}}
    {\color{#2}##1}}%
  \expandafter\newcommand\csname at#1\endcsname{%
    \fcolorbox{#2}{#2}{\color{white}\textsf{\textbf{@#1}}}
    {\color{#2}}}%
  \expandafter\newcommand\csname #1hl\endcsname[2]{%
    \colorbox{#2}{\color{white}\textsf{\textbf{#1}}}\sethlcolor{Azure2}\hl{##2}~%
    \expandafter\ifx\csname commentarrow\endcsname\relax$\leftarrow$\else \commentarrow[#2]\fi~%
    {\color{#2}##1}}%
  \expandafter\newcommand\csname #1st\endcsname[2]{%
    \colorbox{#2}{\color{white}\textsf{\textbf{#1}}}\sout{##2}~%
    \expandafter\ifx\csname commentarrow\endcsname\relax$\leftarrow$\else \commentarrow[#2]\fi~%
    {\color{#2}##1}}%
}
\newcommenter{TODO}{DodgerBlue1}
\newcommenter{rtron}{Green3}

\usepackage{comment}

\usepackage{pdfcomment}

\usepackage{soul}

\usepackage[normalem]{ulem}

\usepackage{csquotes}


%% file: preamble/graphicsTikz.tex
\usepackage{tikz}
\usetikzlibrary{calc}
\usetikzlibrary{matrix}
\usetikzlibrary{chains}
\usetikzlibrary{shapes.geometric}
\usetikzlibrary{arrows.meta}
\usetikzlibrary{decorations.pathreplacing}
\usetikzlibrary{backgrounds}

\tikzset{
  dim above/.style={to path={\pgfextra{
        \pgfinterruptpath
        \draw[>=latex,|->|] let
        \p1=($(\tikztostart)!1.5em!90:(\tikztotarget)$),
        \p2=($(\tikztotarget)!1.5em!-90:(\tikztostart)$)
        in(\p1) -- (\p2) node[pos=.5,sloped,above]{#1};
        \endpgfinterruptpath
      }
    }
  },
  dim double above/.style={to path={\pgfextra{
        \pgfinterruptpath
        \draw[>=latex,|->|] let
        \p1=($(\tikztostart)!3em!90:(\tikztotarget)$),
        \p2=($(\tikztotarget)!3em!-90:(\tikztostart)$)
        in(\p1) -- (\p2) node[pos=.5,sloped,above]{#1};
        \endpgfinterruptpath
      }
    }
  },
  dim below/.style={to path={\pgfextra{
        \pgfinterruptpath
        \draw[>=latex,|->|] let 
        \p1=($(\tikztostart)!-1em!-90:(\tikztotarget)$),
        \p2=($(\tikztotarget)!-1em!90:(\tikztostart)$)
        in (\p1) -- (\p2) node[pos=.5,sloped,below]{#1};
        \endpgfinterruptpath
      }
    }
  },
}

\tikzset{
    right angle quadrant/.code={
        \pgfmathsetmacro\quadranta{{1,1,-1,-1}[#1-1]}     
        \pgfmathsetmacro\quadrantb{{1,-1,-1,1}[#1-1]}},
    right angle quadrant=1, 
    right angle length/.code={\def\rightanglelength{#1}},   
    right angle length=2ex, 
    right angle symbol/.style n args={3}{
        insert path={
            let \p0 = ($(#1)!(#3)!(#2)$) in     
                let \p1 = ($(\p0)!\quadranta*\rightanglelength!(#3)$), 
                \p2 = ($(\p0)!\quadrantb*\rightanglelength!(#2)$) in 
                let \p3 = ($(\p1)+(\p2)-(\p0)$) in  
            (\p1) -- (\p3) -- (\p2)
        }
    }
}

\newcommand{\pgfextractangle}[3]{%
    \pgfmathanglebetweenpoints{\pgfpointanchor{#2}{center}}
                              {\pgfpointanchor{#3}{center}}
    \global\let#1\pgfmathresult  
}

\usetikzlibrary{shapes.arrows}
\newcommand{\commentarrow}[1][Azure4]{\tikz[baseline=-3pt]{\node[shape border uses incircle, fill=#1,rotate=180,single arrow, inner sep=1pt, minimum size=6pt, single arrow head extend=2pt]{};}}

%% file: preamble/robotics.tex
\tikzset{ax/.style={-latex,line width=2pt}}

\tikzset{camera/.style={fill=Sienna1,fill opacity=0.5},%
image plane/.style={draw=RoyalBlue3,line width=2pt}}
